\documentclass[conference]{IEEEtran}
\usepackage{cite}

\ifCLASSINFOpdf
  \usepackage[pdftex]{graphicx}
\else
\fi
\usepackage[cmex10]{amsmath}
\usepackage{fixltx2e}
\usepackage{url}

\usepackage{bm}

\newcommand{\te}[1]{\bm{\mathcal{#1}}}
\newcommand{\ma}[1]{\textbf{#1}}
\newcommand{\ve}[1]{\textbf{#1}}
\newcommand{\sepfigcap}{-0.3cm}

\begin{document}
%
\title{Hot or not? Forecasting cellular network hot spots\\ using sector performance indicators}

\author{
	\IEEEauthorblockN{
		Joan Serr\`a, Ilias Leontiadis, Alexandros Karatzoglou, and Konstantina Papagiannaki
	}
	\IEEEauthorblockA{
		Telef\'onica Research\\
		Pl.~Ernest Lluch i Mart\'in 5\\ 
		08019 Barcelona, Spain\\
		Email: firstname.lastname@telefonica.com
	}
}


\maketitle

\begin{abstract}
To manage and maintain large-scale cellular networks, operators need to know which sectors underperform at any given time. For this purpose, they use the so-called hot spot score, which is the result of a combination of multiple network measurements and reflects the instantaneous overall performance of individual sectors. While operators have a good understanding of the current performance of a network and its overall trend, forecasting the performance of each sector over time is a challenging task, as it is affected by both regular and non-regular events, triggered by human behavior and hardware failures. In this paper, we study the spatio-temporal patterns of the hot spot score and uncover its regularities. Based on our observations, we then explore the possibility to use recent measurements' history to predict future hot spots. To this end, we consider tree-based machine learning models, and study their performance as a function of time, amount of past data, and prediction horizon. Our results indicate that, compared to the best baseline, tree-based models can deliver up to 14\% better forecasts for regular hot spots and 153\% better forecasts for non-regular hot spots. The latter brings strong evidence that, for moderate horizons, forecasts can be made even for sectors exhibiting isolated, non-regular behavior. Overall, our work provides insight into the dynamics of cellular sectors and their predictability. It also paves the way for more proactive network operations with greater forecasting horizons.
\end{abstract}


%
\IEEEpeerreviewmaketitle

\section{Introduction}
\label{sec:introduction}

\begin{figure}[t]
    \centering
    \includegraphics{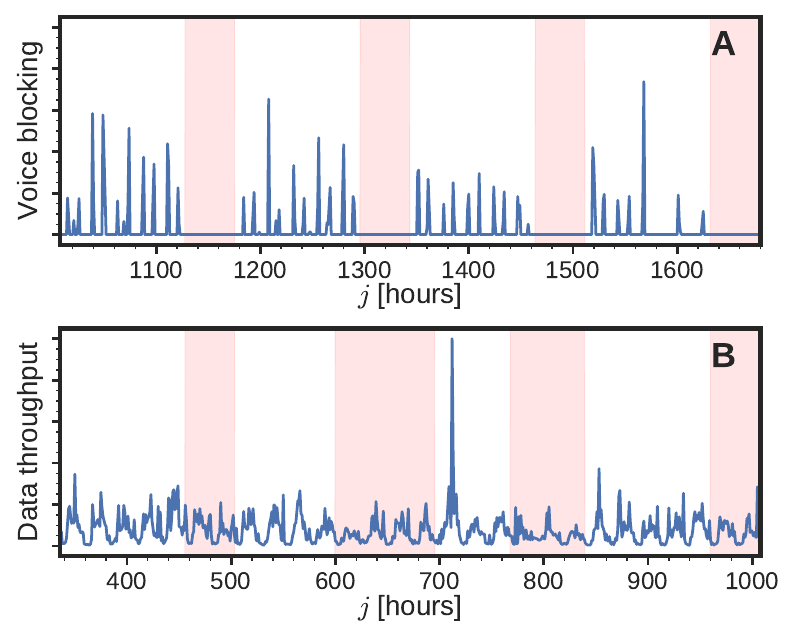}
    \vspace{\sepfigcap}
    \caption{Example of a voice-based (A) and a data-based (B) KPI. The sector in (B) is close to a commercial area, and the strong peak appears in the afternoon of popular shopping day.}
    \label{fig:examplePIs}
\end{figure}

The performance of a cellular network is dynamic; it changes with time, affected by internal and external factors, including hardware and software failures, human behavior, weather conditions, and seasonal changes~\cite{Mahimkar13CONEXT}. Operators make huge investments to accurately monitor and understand performance dynamics, as they are key to cellular networks' management, planning~\cite{planning}, and optimization~\cite{optimization}. Operators rely on a set of equipment measurements that provide detailed performance information: the so-called key performance indicators (KPIs). KPIs provide insights about each cellular sector over a certain time window, typically on the order of minutes or hours~\cite{kpis}. They can be categorized into three broad classes: signaling and coverage monitoring, voice-related measurements, and data-related metrics. Their dynamics can present weekly and workday regularities (Fig.~\ref{fig:examplePIs}A), but also sporadic peaks reflecting non-regular events (Fig.~\ref{fig:examplePIs}B). 

Monitoring the performance of the network can be a complex task, as there are typically hundreds of thousands of sectors providing tens of different KPIs at a constant (almost instantaneous) rate. To facilitate their analysis, operators combine the available KPIs into a single metric that, besides capturing the overall sectors' `health', can be used to rank them based on their performance over time~\cite{nokia_kpis, hotspots, huawei_thresholds}. 
To simplify things further, operators typically only consider `hot spots': sectors that are at the top of the ranking and, thus, clearly underperform. The methodology to combine KPIs into a single metric and the threshold to determine hot spots have been established over the years by vendors and the industry, based on domain knowledge, logical decisions, service level agreements, and controlled experiments~\cite{optimization}.

Understanding the temporal dynamics of hot spots is important. Not only because it provides knowledge about possible root causes of such hot spots~\cite{hotspots}, but also because it brings an intuition on how they will evolve over time. Indeed, by knowing typical hot spot patterns and the regularities of a given sector, one could make a projection into the future and forecast its behavior. Being able to forecast hot spots would provide a number of advantages: (1) as investment plans are finalized weeks in advance, forecasting future demands would allow operators to optimize capex spending~\cite{planning}; (2) short-term forecasting would allow operators to proactively troubleshoot their network~\cite{sprout}; (3) it would also allow to dynamically balance resources as an input to self-organizing networks~\cite{son}.  

In this paper, we study the spatio-temporal dynamics of hot spots. We discover and quantify the most prominent patterns that emerge in a real-world cellular network. Moreover, we develop a methodology to forecast future hot spots. We show evidence on the feasibility of such a task, and analyze the importance of the information contained in the KPIs. To the best of our knowledge, this is the first time that hot spot patterns and hot spot predictability in cellular networks are addressed. A detailed summary of our contributions follows:
\begin{itemize}
\item We consider a large-scale data set of hourly measurements over a period of four months, comprising tens of thousands of sectors, collected for a whole country by a top-tier mobile operator with more than 10~million subscribers (Sec.~\ref{sec:data}). 
\item We mathematically formalize the combination of KPIs into a single metric and the definition of the hot spot score, describing the current practice of cellular network operators (Secs.~\ref{sec:data_notation} and~\ref{sec:data_desc}).
\item We employ a novel data imputation technique based on deep neural networks in order to deal with missing values in the KPI time series (Sec.~\ref{sec:data_missing}).
\item We study the temporal regularities of the hot spot score (Sec.~\ref{sec:data_reg}). We uncover and quantify the most prominent hourly, daily, and weekly patterns.
\item We also study the spatial dynamics of the hot spot score (Sec.~\ref{sec:data_reg}). We show how correlations vanish with distance, but also how very similar behaviors can be found between far away sectors.
\item We provide a forecasting methodology that allows using KPIs to predict the creation and future evolution of hot spots (Secs.~\ref{sec:forecasting_method} and~\ref{sec:forecasting_eval}). We consider four different baseline models, each of them representing a certain forecasting concept using the most basic information of the previous hot spot scores (Sec.~\ref{sec:forecasting_models_baselines}).
\item We develop a set of tree-based forecasting models exploiting all available information (Sec.~\ref{sec:forecasting_models_classifiers}). Such models consider the raw KPIs and hot spot scores, daily percentiles, weekly profiles, and hand-crafted features derived from the previous signals.
\item We perform forecasts in two situations: daily hot spots and emerging persistent hot spots. We evaluate accuracy as a function of time (Sec.~\ref{sec:results_temporal}), prediction horizon (Sec.~\ref{sec:results_horizon}), and amount of considered past information (Sec.~\ref{sec:results_past}). Among others, we show that the time of the forecast does not significantly affect our results, that forecast accuracy reaches a plateau when more than one week of past information is considered, and that tree-based models can outperform the best baseline by 14\% on daily hot spots and by 153\% on emerging hot spots.
\item We analyze the importance of KPIs for forecasting in the two aforementioned situations (Sec.~\ref{sec:results_features}). We show that the most important features are the preceding hot spot scores. However, we also see that usage- and congestion-related KPIs have a non-negligible contribution. Such contribution increases for emerging persistent hot spots.
\end{itemize}
We close the paper by reviewing the related work (Sec.~\ref{sec:relatedwork}) and summarizing our main findings (Sec.~\ref{sec:conclusion}).

\section{Data description}
\label{sec:data}

\subsection{Notation}
\label{sec:data_notation}

We use lowercase italics for single numbers (e.g.,~$q$), lowercase bold for one-dimensional vectors (e.g.,~$\ve{q}$), uppercase bold for two-dimensional matrices (e.g.,~$\ma{Q}$), bold calligraphy for higher-dimensional arrays or tensors (e.g.,~$\te{Q}$), and colons for slices over dimensions (e.g.,~$\te{Q}_{i,1:24,:}$). A quick guide of the used mathematical variables and functions is given in Table~\ref{tab:mathvars}.

\begin{table}[t]
    \caption{List of mathematical variables and functions used in the paper.}
    \centering
    \resizebox{\linewidth}{!}{
    \begin{tabular}{llp{4.5cm}}
        \hline\hline
        Name & Type & Description \\
        \hline
        $\ma{C}$    & Matrix    & Calendar information \\
        $\delta$    & Number  & Temporal integration range \\
        $\Delta$    & Number    & Relative improvement over a model \\
        $F$         & Function  & Forecasting model \\
        $G$         & Function  & Random number generator \\
        h,d,w       & Superindices & Indicators of temporal resolution \\
        $h$         & Number    & Prediction horizon \\
        $H$         & Function  & Heaviside step function \\
        $i,j,k,t$   & Number    & Indices \\
        $\te{K}$    & Tensor    & Key performance indicators \\
        $l$         & Number    & Number of key performance indicators \\
        $\Lambda$   & Number    & Lift over the random model \\
        $\mu$       & Function  & Temporal averaging function \\
        $n$         & Number    & Number of sectors \\
        $m$         & Number    & Length of time series \\
        $\psi$      & Number    & Average precision \\
        $\ma{S}'$   & Matrix    & Sector scores \\
        $\ma{S}$    & Matrix    & Temporally-integrated sector scores \\
        $w$         & Number    & Past information window length \\
        $\te{X}$    & Tensor    & Input to the forecaster (data used for forecasting) \\
        $\ma{Y}$    & Matrix    & Hot spot assignations \\
        $\hat{\ma{Y}}$ & Matrix    & Hot spot forecast \\
        \hline\hline
    \end{tabular}
    }
    \label{tab:mathvars}
\end{table}

\subsection{From key performance indicators to the hot spot score}
\label{sec:data_desc}

We use a set of real-world KPIs collected by a top-tier mobile operator from an entire European country with more than 10~million subscribers\footnote{We explicitly do not show any service performance numbers in the paper for proprietary reasons. However, our observations and inferences are still totally supported by the reported results.}. The employed indicators correspond to 3G sectors, and have been selected based on both internal knowledge and vendors' recommendations. We can group such KPIs into the following classes:  coverage (e.g.,~radio interference, noise level, power characteristics), accessibility (e.g.,~success establishing a voice or data channel, paging success, allocation of high-speed data channels), retainability (e.g.,~fraction of abnormally dropped channels), mobility (e.g.,~handovers' success ratio),  availability and congestion (e.g.,~number of transmission time intervals, number of queued users waiting for a resource, congestion ratios, free channels available). It is important to underline that the data set captures the network at scale, that is, all 3G sectors and all customers observed during the aforementioned period.

In total, we have $l=21$ KPIs for $n$ sectors, where $n$ is on the order of tens of thousands. KPIs are measured hourly for a period of 4~months: from November 30, 2015, to April 3, 2016. This equals to $m_{\text{w}}=18$~weeks, $m_{\text{d}}=126$~days, or $m_{\text{h}}=3024$~hours. We represent these KPIs as a three-dimensional tensor $\te{K}$, whose elements $\te{K}_{i,j,k}$ correspond to the $i$-th sector, $j$-th time sample, and $k$-th indicator. Therefore, $\te{K}$ has a size of $n\times m_{\text{h}}\times l$
.

As mentioned, from the hourly KPIs $\te{K}$, the cellular network operator computes a score $\ma{S}'$ that acts as the main and only measure of `hotness' (or `poor health'), and which can be used to rank sectors based on their performance. A weighted sum of thresholded indicators is used:
\begin{equation}
\ma{S}'_{i,j} = \sum_{k=1}^{d} \bm{\Omega}_{k}H\left(\te{K}_{i,j,k}-\bm{\varepsilon}_k\right) ,
\label{eqn:score}
\end{equation}
where $H$ is the Heaviside step function, and $\bm{\Omega}$ and $\bm{\varepsilon}$ are sets of carefully defined weights and thresholds, respectively. Those have been set and refined over the years, according to the vendor's information and the operator's experience, in order to fulfill management and maintenance needs.

To study the dynamics of the hot spot score, we can focus on different temporal resolutions. To do so, the score $\ma{S}'_{i,j}$ can be integrated over different periods of time, taking averages of subsequent scores. This yields hourly (h), daily (d), and weekly (w) scores, which we can compute from the following expression for a generic integration period $\Gamma$,
\begin{equation}
\ma{S}^{\Gamma}_{i,j} = \mu\left( j\delta_{\Gamma},\delta_{\Gamma},\ma{S}'_{i,:} \right) ,
\label{eqn:s_integration}
\end{equation}
where $\delta_\Gamma$ is the integration length (in hours) for a time period $\Gamma=\{\text{h},\text{d},\text{w}\}$: $\delta_{\text{h}}=1$, $\delta_{\text{d}}=24$, and $\delta_{\text{w}}=168$. The function $\mu(x,y,\ve{z})$ corresponds to a simple average over $\ve{z}$, considering only the time window $y$ preceding $x$:
\begin{equation}
\mu\left(x,y,\ve{z}\right) = \frac{1}{y} \sum_{j=x-y}^{x} \ve{z}_{j} .
\label{eqn:averaging}
\end{equation}

\begin{figure}[t]
    \centering
    \includegraphics{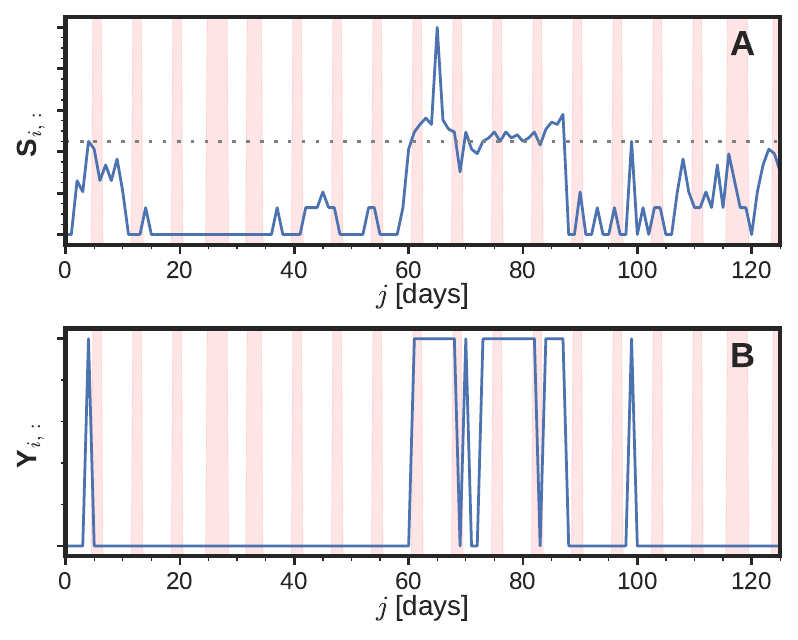}
    \vspace{\sepfigcap}
    \caption{Example of a sector's score $\ma{S}^{\text{d}}_{i,:}$ (A) and its binary hot spot label $\ma{Y}^{\text{d}}_{i,:}$ (B). Red shadowed regions correspond to weekends or holidays.}
    \label{fig:examplescore}
\end{figure}

\begin{figure}[t]
    \centering
    \includegraphics{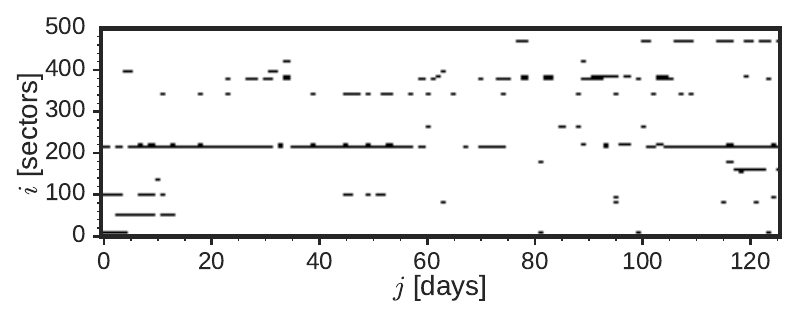}
    \vspace{\sepfigcap}
    \caption{Examples of hot spot labels $\ma{Y}^{\text{d}}$ for 500 randomly selected sectors. Black dots correspond to hot spots.}
    \label{fig:example_y_day}
\end{figure}

\begin{figure}[t]
    \centering
    \includegraphics{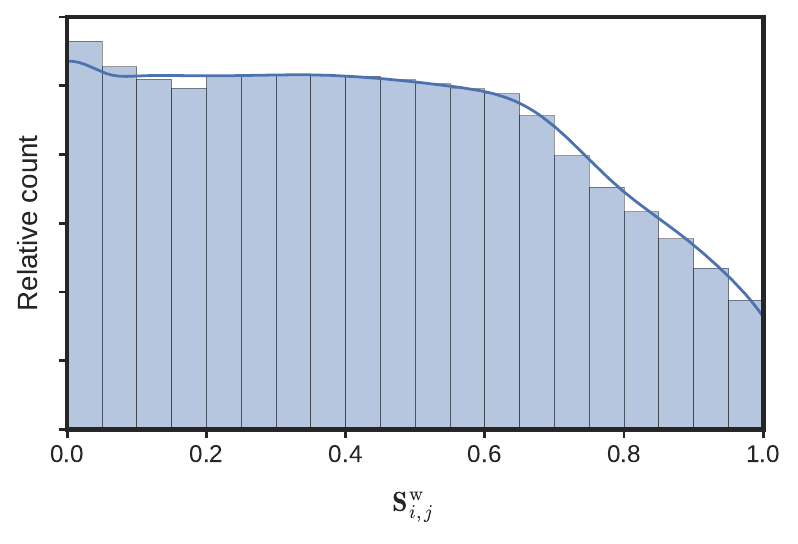}
    \vspace{\sepfigcap}
    \caption{Log histogram of the (re-scaled) hot spot score $\ma{S}^{\text{w}}$. Notice the presence of a natural threshold at $\ma{S}^{\text{w}}\approx 0.6$.}
    \label{fig:scoredist}
\end{figure}

In order to further simplify the analysis of underperforming sectors, the operator performs an explicit hot spot label designation. A sector $i$ is defined as a hot spot at a given time step $j$ if $\ma{S}_{i,j}$ is above an empirically chosen threshold $\epsilon$ (Fig.~\ref{fig:examplescore}):
\begin{equation}
\ma{Y}_{i,j} = H(\ma{S}_{i,j}-\epsilon) .
\label{eqn:bigthreshold}
\end{equation}
The result is a binary signal $\ma{Y}_{i,j}$ denoting whether a sector $i$ is a hot spot or not at a given time frame $j$ (Fig.~\ref{fig:example_y_day}). The choice of the appropriate threshold is based on the operator's best practice, but can also be partially justified from the empirical distribution of $\ma{S}$ (Fig.~\ref{fig:scoredist}).

Besides KPIs $\te{K}$, scores $\ma{S}$, and hot spot labels $\ma{Y}$, we also have a timestamp that can give us enriched calendar information. In our case, this consists of 5 different vectors reflecting (1) the hour of the day, (2) the day of the week, (3) the day of the month, (4) if it is a weekend, and (5) if it is a holiday. We concatenate these 5 vectors to form an $m_{\text{h}}\times 5$ matrix $\ma{C}$, performing brute-force upsampling for all signals except (1) in order to obtain hourly values
.

\subsection{Dealing with missing values}
\label{sec:data_missing}

Unfortunately, KPIs are not always present for every sector, hour, and indicator. This can be due to multiple reasons, including the cases where the site was offline, the backbone was congested (priority is given to customer's data), the collection server was congested, or the probes have malfunctioned. In general, we find missing values for a specific sector, hour, and KPI ($\te{K}_{i,j,k}$), for slices over indicators of a specific sector and hour ($\te{K}_{i,j,:}$), and for slices over time and indicator for a given sector ($\te{K}_{i,j:j+t,:}$). 

In order to deal with missing values, we proceed in two steps. Firstly, we perform a sector filtering operation, discarding a sector $i$ if it has more than 50\% of missing values in one or more weeks, that is, if
\begin{equation*}
\frac{1}{\delta_{\text{w}}} \sum_{t=j}^{j+\delta_{\text{w}}} \frac{1}{l} \sum_{k=1}^{l} M(\te{K}_{i,t,k}) > 0.5
\end{equation*}
for any $1\leq j \leq m_{\text{h}}-\delta_{\text{w}}$, where $M(x)=1$ if the value of $x$ is missing and 0 otherwise. This filtering operation discards around 10\% of the sectors, yielding $n$ still in the range of tens of thousands. The number of missing values after filtering represents around 4\% of our data.

\begin{figure}[!t]
    \centering
    \includegraphics{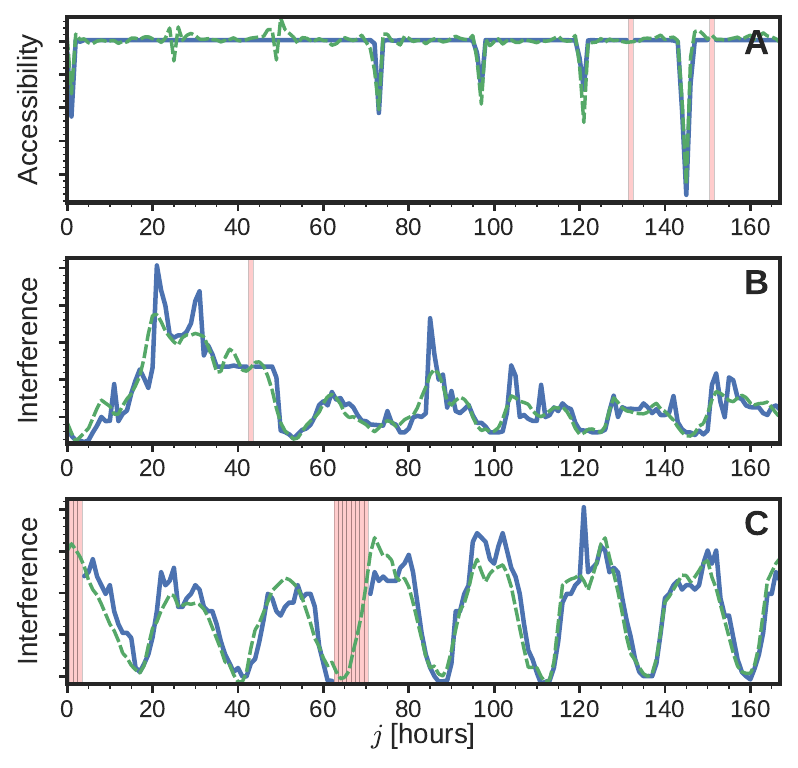}
    \vspace{\sepfigcap}
    \caption{Examples of KPIs (solid blue lines) and their learned reconstructions (dashed green lines). Only missing values (red patches) are replaced by the reconstruction.}
    \label{fig:missvals}
\end{figure}

With the remaining sectors, we perform missing value imputation on a weekly basis. For that, we use a stacked denoising autoencoder~\cite{Vincent11JMLR}. An autoencoder is a multi-layer neural network that is trained to attempt to replicate its input at its output, typically performing an implicit dimensionality reduction in the inner layers so that it does not learn the identity function but some structural features of the data~\cite{Goodfellow16BOOK}. A denoising autoencoder goes one step further and tries to reconstruct a corrupted input signal to its original, uncorrupted version. This schema can be directly used for missing value imputation: we can input signals with (artificially introduced) missing values and train the autoencoder to minimize the error between the reconstructed and the original signals (Fig.~\ref{fig:missvals}).

We consider a four-layer encoder and its symmetric decoder. The encoder layers are dense layers of half of the size of their input, with parametric rectified linear units~\cite{He15ICCV} as activation functions. We form batches of 128 randomly chosen subsequences, each of them corresponding to a one week slice over all indicators, 
$\te{K}_{i,\delta_{\text{w}}(j-1)+1:\delta_{\text{w}}j,:}$, 
using $i=G(1,n)$ and  $j=G(1,m_{\text{w}})$, where $G(x,y)$ is a uniform random integer generator such that $x\leq G(x,y)\leq y$.

At the encoder input, we substitute the missing values of every subsequence by the first available previous time sample. The remaining non-missing values up to half of the slice size are also substituted by the first available previous time sample. As loss function we use the mean squared error between the real input and the reconstruction, considering only the originally non-missing values. After some experimentation, we train the autoencoder using RMSprop~\cite{Tieleman12COURSERA} for 1000 epochs of $nm_{\text{w}}/128$ batches each, using a learning rate of $10^{-4}$ and a smoothing factor of 0.99. Z-normalization is applied for each KPI before imputation, and the original offsets and scales are restored afterwards.

\begin{figure*}[t]
    \centering
    \includegraphics{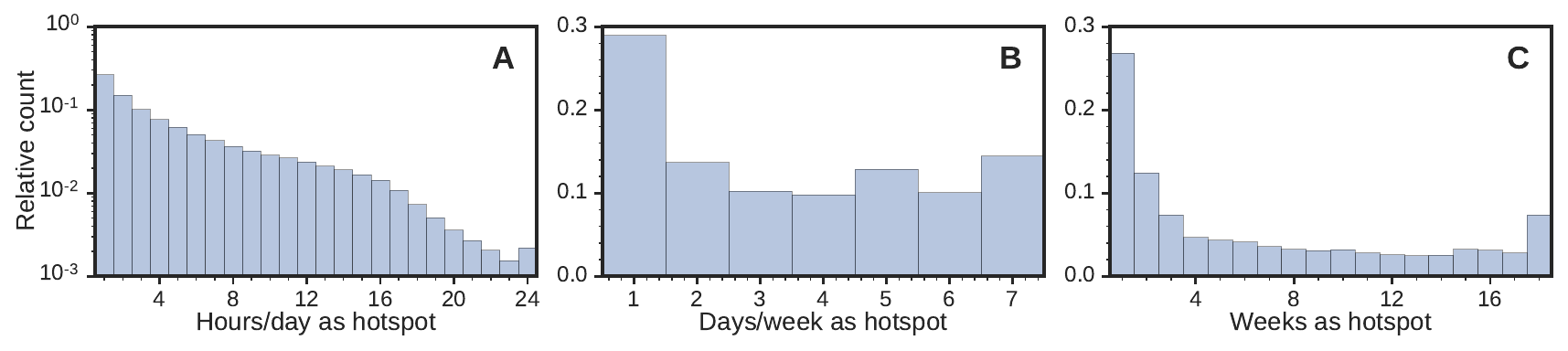}
    \vspace{\sepfigcap}
    \caption{Normalized histograms of the hours per day as hot spot (A), days per week as hot spot (B), and number of weeks as hot spot (C). Notice the logarithmic vertical axis in (A). }
    \label{fig:basic1}
\end{figure*}

\section{Hot spot dynamics}
\label{sec:data_reg}

Before performing the actual forecasting, we want to justify its feasibility and learn about the data by exploring its dynamics. In the previous sections, we have already shown examples of voice-based and data-based KPIs (Fig.~\ref{fig:examplePIs}), of a sector's score and hot spot labels (Figs.~\ref{fig:examplescore} and~\ref{fig:example_y_day}), and of the empirical distribution of hourly sector scores (Fig.~\ref{fig:scoredist}). We now focus on the hot spot labels $\ma{Y}^{\text{h}}$ and $\ma{Y}^{\text{d}}$, and show some of their aggregated statistics. 

First of all, we can study the number of hours that a given sector stays as a hot spot (Fig.~\ref{fig:basic1}A). With that, we empirically find a threshold around 16~hours, which intuitively matches with an 8~hour sleeping pattern. We can do the same on a larger temporal scale, counting how many days a sector has been a hot spot (Fig.~\ref{fig:basic1}B). In this case, we see that the most prominent peak corresponds to 1~day, which means that many sectors become hot only once in a week. The other prominent peaks, although smaller than the 1-day peak, are at 2, 5, and 7~days. Peaks of 2 and 5~days match with weekends and working days, respectively, and peaks of 7~days indicate that there are sectors which are hot for the entire week or perhaps more than one week. In fact, if we go to an even larger temporal resolution and count the number of weeks as a hot spot, we see that some fraction of the available sectors are hot for the entire considered period of $m_{\text{w}}=18$~weeks (Fig.~\ref{fig:basic1}C). Besides this, we observe that the most common value for number of weeks as a hot spot is below 4.

\begin{figure}[t]
    \centering
    \includegraphics{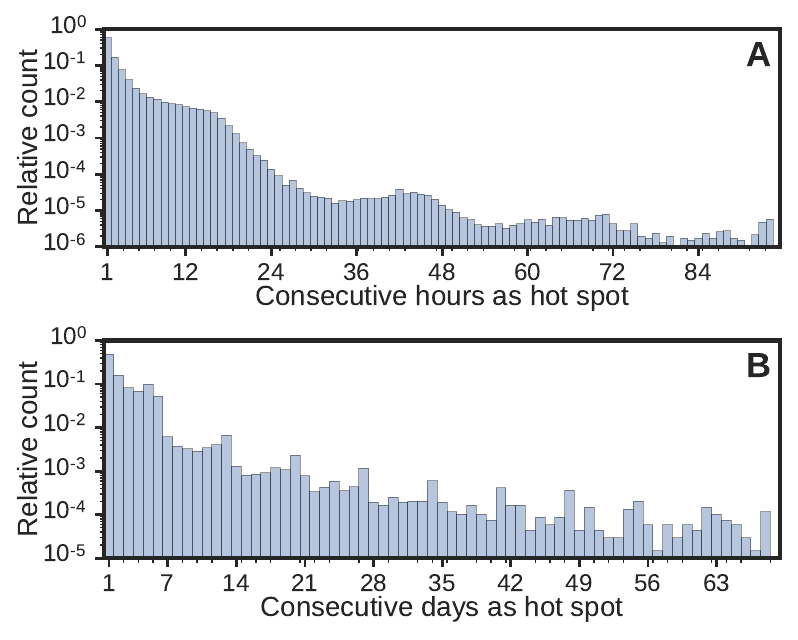}
    \vspace{\sepfigcap}
    \caption{Normalized histograms of consecutive hours as hot spot (A) and consecutive days as hot spot (B). Notice the logarithmic vertical axes.}
    \label{fig:basic2}
\end{figure}

We can get similar insights if we count the number of consecutive time periods that a sector has been a hot spot. If we study the number of consecutive hours as a hot spot, we find the aforementioned peak at 16~hours, but also smaller peaks around 40 (24+16) and 64 (48+16) hours (Fig.~\ref{fig:basic2}A). If we study the number of consecutive days as a hot spot, we can also discover some clear patterns (Fig.~\ref{fig:basic2}B). We observe that the dominant value is 1, indicating the presence of sporadic single-day hot spots. Besides, we also observe the aforementioned peaks at 2 and 5~days. However, we can now see that bursts of 6 consecutive days are more prominent than larger ones. Interestingly, besides peaks at multiples of 7, we now also find peaks at multiples of 7 plus 6 days (13, 20, 27, and so on). We hypothesize that these are sectors that are relatively busy from Monday to Saturday that, in addition, at some times, also get busy on Sunday, hence producing a $7x+6$ pattern. 

A further look into weekly patterns reveals that, in general, workday patterns tend to be more prominent than weekend ones (Table~\ref{tab:patterns}). The most noticeable workday patterns occupy positions 2, 3, and 4, while purely weekend-based patterns only appear at positions 6, 10, and 20. We further analyzed the temporal consistency of these weekly patterns and computed the correlation between the average week pattern versus all the patterns for a given sector. We obtained an average correlation of 0.6, with 5, 25, 50, 75, and 95 percentiles at --0.09, 0.41, 0.68, 0.88, and 1, respectively. Thus, we see that the temporal consistency across sectors is relatively high. This suggests that a forecasting model exploiting such consistency could possibly perform reliable predictions.

\begin{table}[t]
    \caption{Top 20 hot spot weekly patterns and their relative counts. Capital letters correspond to the day of the hot spot and a hyphen corresponds to no hot spot. Counts exclude the `never-hot' pattern (rank 1) for confidentiality reasons and are normalized afterwards. Note that, for 7~days, there are a total of 127 possible different patterns.}
    \centering
    \resizebox{\linewidth}{!}{
    \begin{tabular}{c|c|c}
        \hline\hline
        Rank & Pattern & Count~[\%] \\ 
        \hline
1 	& \texttt{~ -~ -~ -~ -~ -~ -~ - ~} & \\
2 	& \texttt{~ M~ T~ W~ T~ F~ S~ S ~} &	14.4 \\
3 	& \texttt{~ M~ T~ W~ T~ F~ -~ - ~} &	8.5 \\
4 	& \texttt{~ M~ T~ W~ T~ F~ S~ - ~} &	7.2 \\
5 	& \texttt{~ -~ -~ -~ -~ F~ -~ - ~} &	5.4 \\
6 	& \texttt{~ -~ -~ -~ -~ -~ S~ - ~} &	4.7 \\
7 	& \texttt{~ M~ -~ -~ -~ -~ -~ - ~} &	4.1 \\
8 	& \texttt{~ -~ T~ -~ -~ -~ -~ - ~} &	4.1 \\
9 	& \texttt{~ -~ -~ -~ T~ -~ -~ - ~} &	3.9 \\
10 	& \texttt{~ -~ -~ -~ -~ -~ -~ S ~} &	3.5 \\
11 	& \texttt{~ -~ -~ W~ -~ -~ -~ - ~} &	3.2 \\
12 	& \texttt{~ -~ T~ W~ T~ F~ -~ - ~} &	2.4 \\
13 	& \texttt{~ M~ T~ W~ T~ -~ -~ - ~} &	2.3 \\
14 	& \texttt{~ -~ -~ -~ T~ F~ -~ - ~} &	1.7 \\
15 	& \texttt{~ M~ T~ -~ -~ -~ -~ - ~} &	1.6 \\
16 	& \texttt{~ -~ -~ -~ -~ F~ S~ - ~} &	1.5 \\
17 	& \texttt{~ M~ T~ W~ -~ -~ -~ - ~} &	1.4 \\
18 	& \texttt{~ -~ -~ W~ T~ F~ -~ - ~} &	1.4 \\
19 	& \texttt{~ -~ -~ W~ T~ -~ -~ - ~} &	1.3 \\
20 	& \texttt{~ -~ -~ -~ -~ -~ S~ S ~} &	1.3 \\
        \hline\hline
    \end{tabular}
    }
    \label{tab:patterns}
\end{table}

\begin{figure}[t]
    \centering
    \includegraphics{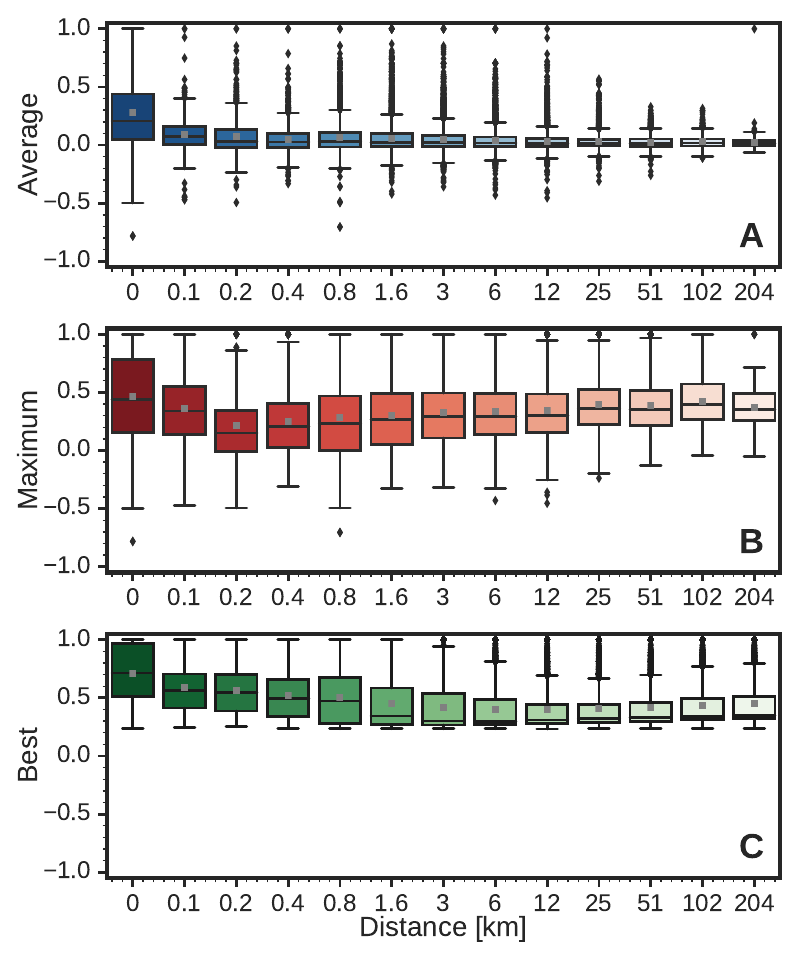}
    \vspace{\sepfigcap}
    \caption{Hot spot temporal sequence correlation between sectors as a function of the physical distance separating them: sector average (A), sector maximum (B), and best possibility (C).}
    \label{fig:spatial}
\end{figure}

We now switch and explore spatial regularities. For that, we study the correlations between the evolving sectors' hotness and the physical separation between them, using the hot spot labels $\ma{Y}^{\text{h}}$. For each sector, we query the spatially closest 500 sectors and compute the Pearson's correlation coefficient between the time series of the former and all the time series of the latter. That is, between $\ma{Y}^h_{i,:}$ and $\ma{Y}^h_{j_i,:}$, where $j_i$ belongs to the set of 500 spatially closer sectors to sector $i$. We then distribute all $500n$ correlation values across logarithmically-spaced spatial buckets and take a per-sector average. This gives us an intuition of the average similarity of hot spot sequences as a function of the sectors' distance (Fig.~\ref{fig:spatial}A). We observe that sectors corresponding to the same tower (at a distance of 0\,km) present the highest correlations. This is expected, as these sectors may cover the same area and they are typically handled by the same base station equipment. Thus, if there is a failure, it can affect all the  sectors of the site. We also observe that the median (across sectors) of the average correlation rapidly decreases with distance, being almost constant at 0 for distances above 100\,m. However, we find that the upper bars and the number and range of the upper outliers decrease more slowly. This indicates that there exist some spatial correlations that eventually vanish with distance. However, since the median is close to 0, these occur for less than half of the spatially closest neighbors.

If instead of taking a per-sector average we take the per-sector maximum, that is, we only focus on the most correlated sector inside a spatial bucket for every sector, we see that an interesting phenomenon arises (Fig.~\ref{fig:spatial}B): the median bucket correlation slightly increases with distance, and the upper range delimiter is almost always above 0.9. This could be an indication that one can find highly-correlated sector behaviors independently of the physical distance that separates them. Intuitively, this makes sense if we think of spatial distributions of land use: areas with similar usage do not necessarily need to be spatially closer, and urban share is one of those usages that can be scattered across geography (cf.~\cite{Chakir12EE,LaRosa13UP}). Thus, in essence, we can expect very similar hot spot sequence behaviors even if the sector locations are very far apart. 

To confirm this hypothesis, we repeat the previous experiment but with some changes. This time, for each sector, we search for the most correlated 100 sectors, independently of the physical distance. We then distribute all $100n$ correlation values across logarithmic-spaced spatial buckets, and take the per-sector maximum. This intuitively corresponds to finding the most similar sector behavior in a given spatial bucket (Fig.~\ref{fig:spatial}C). We observe that most correlations are around 0.5 for all buckets, and that typically a large fraction of the sectors present a correlation above 0.5 at any given spatial bucket. Thus, we can conclude that very similar hot spot sequences are observed independently of the sectors' separation. We will use this knowledge in our forecasting algorithm by preventing it from any spatial constraint. With this, we aim to capture similar hot spot temporal behaviors that occur at both close and far away sectors. Overall, given all the temporal regularities we have seen, we are moderately optimistic about the possibility of performing hot spot forecasting using temporal KPI sequences and hot spot labels.

\section{Hot spot forecasting}
\label{sec:forecasting}

\subsection{Methodology}
\label{sec:forecasting_method}

As mentioned, we want exploit the KPI information to predict if a sector is going to be a hot spot or not. As the main objective for predictions is intervention, and such interventions cannot be typically made `on-the-hour' by operators, we will work with daily resolution. Therefore, our target variable will be the binary labels $\ma{Y}^{\text{d}}$, which correspond to the notion of `being a hot spot' at a certain day (Sec.~\ref{sec:data_desc}). In addition to this target, we will also consider the `become a hot spot' target. The consideration of this second target will allow us to go one step further and study the behavior of our methods on an important subset of the sectors: the ones that were not hot spots for a period of time, but became hot spots consistently for the next few days. To obtain the `become a hot spot' target, we take weekly averages of the daily scores before and after any given day $j$, threshold them, and discard consecutive activations. Following the introduced notation (Eqs.~\ref{eqn:score}--\ref{eqn:bigthreshold}), the `become a hot spot' label can be formalized by
\begin{equation*}
\begin{split}
\ma{Y}_{i,j} = ~ & H\left( \mu\left(j,\delta_{\text{w}},\ma{S}'_{i,:}\right) - \epsilon \right) \\
    & \left(1-H\left( \mu\left(j+\delta_{\text{w}},\delta_{\text{w}},\ma{S}'_{i,:}\right) - \epsilon \right)\right) \\
    & \left(1- H\left(\ma{S}_{i,j} - \epsilon \right)\right) H\left(\ma{S}_{i,j+1} - \epsilon \right) ,
\end{split}
\end{equation*}
where the first two terms relate to weekly averages of the hot spot score and the last two terms focus on discarding consecutive activations.

Notice that, in general, the fact that the target variable has a certain temporal resolution does not necessarily imply that the input variables need to be at the same resolution. In fact, we can gain information from a higher temporal resolution with respect to the target variable. For instance, in our case, hourly KPI fluctuations or trends could become an important indicator of whether some sector will become hot or not on a given day. Therefore, we will use the hourly resolution of our KPIs, $\te{K}^{\text{h}}$, as input to our models. In addition, given that modern machine learning techniques allow to perform feature selection at training time without overfitting, we will consider all available information we have besides KPIs $\te{K}$: calendar information $\ma{C}$, scores $\ma{S}$, and the values of the previous hot spot labels present in $\ma{Y}$. We combine all these variables into one tensor
\begin{equation}
\begin{split}
\te{X} = ~ & [ ~ \te{K} ~ ||_3 ~ R_1(n,\ma{C}) ~ ||_3 ~ \ma{S}^{\text{h}} ~ ||_3 \\
   & U_1(\delta_{\text{d}},\ma{S}^{\text{d}}) ~ ||_3 ~ U_1(\delta_{\text{w}},\ma{S}^{\text{w}}) ~ ||_3 ~ U_1(\delta_{\text{d}},\ma{Y}^{\text{d}}) ~ ] ,
\end{split}
\label{eqn:combine}
\end{equation}
where $||_3$ denotes tensor concatenation along the third dimension, $R_1(k,\ma{X})$ is a function that repeats $k$ times the content of $\ma{X}$ and creates a new first dimension with these repetitions, and $U_1(k,\ma{X})$ is a function that performs brute-force upsampling of $\ma{X}$ along the first dimension by a factor of $k$. In total, $\te{X}$ has a size of $n\times m_{\text{h}}\times (l+5+3+1)$.

Having seen how to combine the available input variables, we can now formulate our forecasting task. Let $t$ be the current day and $h\geq 1$ our desired prediction horizon, that is, how many days into the future we want to perform the forecast. As we cannot input all past information to our algorithm, we also need to define a temporal window $w\geq 1$ from which we are going to get our input variables. The introduction of this temporal window has a threefold motivation. Firstly, the more into the past our input values lie, the less likely it is to help in a reliable forecast. This intuitive statement has been formally confirmed in the literature for uncountable scenarios and information sources~\cite{Kantz04BOOK}. Secondly, machine learning algorithms can suffer with an unreasonably high dimensionality, which can hamper their accuracy and interpretability~\cite{Hughes68TIT}. Thirdly, we do not dispose of infinite data, and we need to save part of it for testing/evaluation purposes~\cite{Hyndman13BOOK}.

With $t$, $h$, and $w$, we can now formulate our prediction variable
\begin{equation}
\hat{\ma{Y}}_{i,t+h} = F\left( \te{X}_{i,t-w:t,:} \right),
\label{eqn:prediction}
\end{equation}
where $F$ represents a forecasting model that outputs $\hat{\ma{Y}}_{i,t+h}$, the probability of being a hot spot at horizon $h$. The input of the model is $\te{X}_{i,t-w:t,:}$, the slice over sector $i$, the previous temporal window $w$, and all feature variables (including all the variables in the third dimension of the tensor $\te{X}$, Eq.~\ref{eqn:combine}). Note that, for ease of notation, from now on we will use day resolution for tensor indices. Therefore, in Eq.~\ref{eqn:prediction}, the slice $t-w:t$ (in days) implies $t-24w:t$ in hours. 

At training time, we however only dispose of data until $t$. Therefore, the training of our model $F$ will need to be done with the corresponding $h$ times delayed slice $\te{X}_{i,t-h-w:t-h,:}$. Moreover, we will use our binary hot spot label $\ma{Y}_{i,t}$. With that,
\begin{equation}
\ma{Y}_{i,t} = F\left( \te{X}_{i,t-h-w:t-h,:} \right).
\label{eqn:training}
\end{equation}
Therefore, our setting corresponds to a typical classification task where models are trained with binary inputs and provide real-valued probability outputs~\cite{Hastie09BOOK}.

\subsection{Evaluation measures}
\label{sec:forecasting_eval}

Evaluation is performed with probabilities $\hat{\ma{Y}}_{:,t+h}$ and binary variables $\ma{Y}_{:,t+h}$. In essence, we want to obtain a ranking of the available sectors using probabilities $\hat{\ma{Y}}_{:,t+h}$ (largest values first), where topmost sectors correspond to the true hot spot sectors $i$ that have labels $\ma{Y}_{i,t+h}=1$. This corresponds to a classical information retrieval task where relevant documents to a given query need to be ranked first in the provided answer~\cite{Manning08BOOK}. Thus, we can use classical and well-known information retrieval measures. In particular, we work with precision-recall curves and the average precision measure~\cite{Manning08BOOK}. However, as both are sensitive to the number of positive instances, and therefore could give a rough estimate of the total number of hot spots, which we cannot reveal, we will here report lift values relative to random performance~\cite{Hastie09BOOK}. In fact, because of such sensitivity, both measures are always assessed with respect to random performance~\cite{Manning08BOOK}. 

The lift of a model $F^i$ over the random model $F^0$ is defined as
$\Lambda^i = \psi(F^i)/\psi(F^0)$,
where $\psi(X)$ represents the average precision of model $X$. Note that a lift $\Lambda^i\approx 1$ indicates close to random performance, and that a lift of $\Lambda^i$ implies a performance that is $\Lambda^i$ times better than random. We will compute lift values for every $t$, $h$, $w$, and model combination, and report results based on different statistics across subsets of the four variables. Apart from lift from random, we will also report differences between any two models $F^i$ and $F^j$ in a relative way, using ratio percentages as computed by
$\Delta^{ij} = 100(\Lambda^j/\Lambda^i-1)$.

\subsection{Forecast models -- Baselines}
\label{sec:forecasting_models_baselines}

\vspace{0.0cm}\noindent\textbf{Random model:} 
We measure random performance using  
$\hat{\ma{Y}}_{i,t+h} = G(0,1)$,
where $G(x,y)$ is a uniform random real number generator such that $x\leq G(x,y) \leq y$. This will be our $F^0$ model. As mentioned, the results for this model will give us an indication of chance level.

\vspace{0.2cm}\noindent\textbf{Persist model:} 
Another trivial model to use as baseline in any forecasting task is the persistence model~\cite{Hyndman13BOOK}. This model will just exploit the current value of the ground truth variable to perform a forecast:
$\hat{\ma{Y}}_{i,t+h} = \ma{Y}_{i,t}$.
This model can actually provide very good forecasts in situations where the variable $\ma{Y}$ comes in bursts or does not change much with time~\cite{Kantz04BOOK}. The results for this model thus give us an indication of how persistent is the signal we are forecasting in the short term.

\vspace{0.2cm}\noindent\textbf{Average model:} 
Going a little bit farther, we can decide to exploit the more fine-grained value of $\ma{S}$ instead of $\ma{Y}$. We can then define the average future score
$\hat{\ma{Y}}_{i,t+h} = \mu\left(t,w,\ma{S}_{i,:}\right)$,
where $\mu$ represents an average of the preceding values (Eq.~\ref{eqn:averaging}). Note that the forecast $\hat{\ma{Y}}_{:,t+h}$ is not a probability, but can nonetheless be used for ranking hot sectors and, therefore, for computing $\psi$. The results for this model give us an indication of how stationary is the signal we are forecasting in the mid term.

\vspace{0.2cm}\noindent\textbf{Trend model:} 
Besides the previous static models, we can also enhance the Average model and consider a more dynamic one by aggregating a future projection of the current trend in the hot spot score $\ma{S}$:
\begin{equation*}
\hat{\ma{Y}}_{i,t+h} \! = \! \mu\left(t,w,\ma{S}_{i,:} \right) + \frac{ \mu\left(t,\frac{w}{2},\ma{S}_{i,:} \right) - \mu\left(t\!\!-\!\!\frac{w}{2},\frac{w}{2},\ma{S}_{i,:} \right) }{ \frac{w}{2} } .
\end{equation*}
Again, the output values are not probabilities, but can be used for ranking and computing $\psi$. The results for this model give us an indication of how trivial is the signal evolution in the short/mid term.

\subsection{Forecast models -- Classifiers}
\label{sec:forecasting_models_classifiers}

\vspace{0.0cm}\noindent\textbf{Tree model:} 
Not all machine learning models are able to deal with a large number of instances $n$ and heterogeneous features of different scales and distributions, which corresponds to the situation we face with KPIs. One of the models that fulfills such requirements is the classification and regression tree~\cite{Breiman84BOOK}. Classification trees could be regarded as the `off-the-shelf' procedure for classification, as they are invariant under scaling and other feature transformations, robust to the inclusion of irrelevant features, and produce interpretable models~\cite{Hastie09BOOK}. They perform a recursive partitioning of the training set until, ideally, elements of the same label or class predominate in a partition. There are several stopping criteria for the partitioning, as well as metrics to evaluate the best split point and feature~\cite{Hastie09BOOK}. 

In our study, we employ the implementation in the \texttt{scikit-learn} package~\cite{Pedregosa11JMLR} (version 0.17.0). We use the Gini coefficient~\cite{Hastie09BOOK} as the split metric, and evaluate a random subset of 80\% of the features at every partition of the tree. Sample weights are balanced by multiplying by the inverse of the class frequency, and a 2\% of the total weight is used as a criterion to stop the partitioning of the tree. 

\vspace{0.2cm}\noindent\textbf{Random forest:} 
It is well known however that classification and regression trees present a series of limitations, including local optimality effects and poor generalization capabilities~\cite{Hastie09BOOK}. To overcome some of those limitations, an ensemble of trees can be built, and predictions can be aggregated to reduce local optimality and improve generalization. A classical approach to that is random forests~\cite{Ho95ICDAR}. Random forests train a number of trees on randomized subsets of the available instances, using also a randomized subset of the features at each tree partition~\cite{Breiman01ML}. Predictions can then be performed by, for instance, aggregating the class probabilities computed for each tree in a so-called bootstrap aggregating or bagging schema~\cite{Breiman96ML}.

We again use the implementation in \texttt{scikit-learn}, with the same balancing strategy and split metric as with the Tree model. However, at every partition, we now only evaluate a random subset of the features whose size cannot exceed the square root of the total number of input features~\cite{Breiman01ML}. In addition, we build much deeper trees by considering 0.02\% of the total weight as the criterion to stop the partitioning of a tree. It is demonstrated that random forests can use deep decision trees without the risk of overfitting to the training data~\cite{Hastie09BOOK}. Random forests lose a little bit of the interpretability of trees but, nonetheless, can be used to derive feature importances in an intuitive way~\cite{Breiman01ML}. The input features for the random forest is the raw slice $\te{X}_{:,j-w:j,:}$, where $j=t-h$ for training and $j=t$ for forecasting (Eqs.~\ref{eqn:training} and~\ref{eqn:prediction}, respectively). We denote this model by RF-R.

\vspace{0.2cm}\noindent\textbf{Random forest with percentile features:} 
Despite the generalization and feature selection capabilities of a machine learning algorithm, it is well known that the total number~\cite{Hughes68TIT} and the quality of such features is crucial to obtain accurate predictions~\cite{Hastie09BOOK}. Therefore, to improve both aspects in our forecasting task, we decide to summarize the content of the data before being input to the random forest classifier. In our first attempt to achieve that, we construct $w$ daily percentile estimators from input time series $\te{X}_{:,j-w:j,:}$, where $j=t-h$ for training and $j=t$ for forecasting. We use the 5, 25, 50, 75, and 95 percentiles of every day $j$ of each $\te{X}_{i,j-w:j,k}$. Note that percentiles are computed using every 24 samples in the second dimension, thus we reduce the dimensionality of every slice from $24w$ to $5w$. Note furthermore that this feature set implicitly includes the Persistence and Average models described above. We denote this model by RF-F1.

\vspace{0.2cm}\noindent\textbf{Random forest with hand-crafted features:} 
We also experiment with a set of hand-crafted features computed from the slice $\te{X}_{:,j-w:j,:}$, where again $j=t-h$ for training and $j=t$ for forecasting. We consider the mean, standard deviation, minimum, and maximum of the whole time series $\te{X}_{i,j-w:j,k}$, its first half, $\te{X}_{i,j-w:j-w/2,k}$, and its second half, $\te{X}_{i,j-w/2:j,k}$. We also consider the differences between the quantities derived for the two halves $\te{X}_{i,j-w/2:j,k}$ and $\te{X}_{i,j-w:j-w/2,k}$. In addition, we compute average day and week profiles found in the whole time series $\te{X}_{i,j-w:j,k}$, and a set of differences between some of the obtained average components. We also compute what we call `extreme' day and week profiles, by accumulating minimum and maximum values on an hourly and daily basis, respectively. Finally, we include the raw values for the 24 hours of the last day, $\te{X}_{i,j,k}$, plus the mean and standard deviation of those. Note that this feature set implicitly includes the Persistence, Average, and Trend models described above. We denote this model by RF-F2.

\subsection{Tasks' summary}

We want to assess the performance of each of the aforementioned models as a function of $t$, $h$, and $w$. With $t$, we want to study the performance variability with respect to time. With $h$, we want to study how far into the future we can perform a reliable prediction. With $w$, we want to study how much past information we need to consider in order to perform such predictions. In total, we have a number of combinations between these four main variables of interest (Table~\ref{tab:valcomb}).

\begin{table}[t]
    \centering
    \caption{Considered values for model, time step $t$, prediction horizon $h$, and amount of past $w$.}
    \resizebox{\linewidth}{!}{
    \begin{tabular}{lp{7.5cm}}
        \hline\hline
        Variable & Considered values \\ 
        \hline
        Model   & $\{$Random, Persist, Average, Trend, Tree, RF-R, RF-F1, RF-F2$\}$ \\
        $t$     & $\{$52, 53, 54, $\dots$ 87$\}$\\
        $h$     & $\{$1, 2, 3, 4, 5, 7, 8, 10, 12, 14, 16, 19, 22, 26, 29$\}$\\
        $w$     & $\{$1, 2, 3, 5, 7, 10, 14, 21$\}$ \\
        \hline\hline
    \end{tabular}
    }
    \label{tab:valcomb}
\end{table}

\section{Forecasting results}
\label{sec:results}

\subsection{Temporal stability}
\label{sec:results_temporal}

The first aspect we consider is the dependence of the results with respect to the day of our analysis. More specifically, given a combination of model, $h$, and $w$, we assess whether there is any significant performance variation in $t$. To do so, we perform a statistical hypothesis test to quantify differences in the performance distributions obtained for different time periods. First, we create two equal-sized splits of the variable $t$: $t\in[52,69]$ and $t\in[70,87]$. Next, we take the average precision values $\psi$ corresponding to the two splits and perform a two-sample Kolmogorov-Smirnov test~\cite{Hollander99BOOK}. The Kolmogorov-Smirnov test is a non-parametric test for the equality of continuous, one-dimensional probability distributions that is sensitive to differences in both location and shape of the empirical cumulative distribution. It can be used to compare a sample with a reference probability distribution or, as in our case, to compare two samples. Thus, the $p$-values of the test can be used to assess to the null hypothesis that the distribution of the two samples is the same. One typically rejects the null hypothesis when $p$ has a very small value, usually using $p<0.01$ or $p<0.05$.

We independently compute the $p$-values between the two splits for all possible combinations of model, $h$, and $w$. We find that none of such $p$-values is under 0.01, and that only 1.1\% of them is below 0.05. This is strong evidence that the two distributions of average precision values $\psi$ do not present a significant difference, and the evidence becomes even stronger if we consider the effect of multiple comparisons~\cite{Miller81BOOK}. A similar analysis was conducted with further splits based on the day of the week, and also with the `become a hot spot' labels, obtaining almost the same results. Thus, in the remaining, we can safely assume that there is no important variability of the results with respect to $t$.

\subsection{Prediction horizon}
\label{sec:results_horizon}

\begin{figure}[t]
    \centering
    \includegraphics{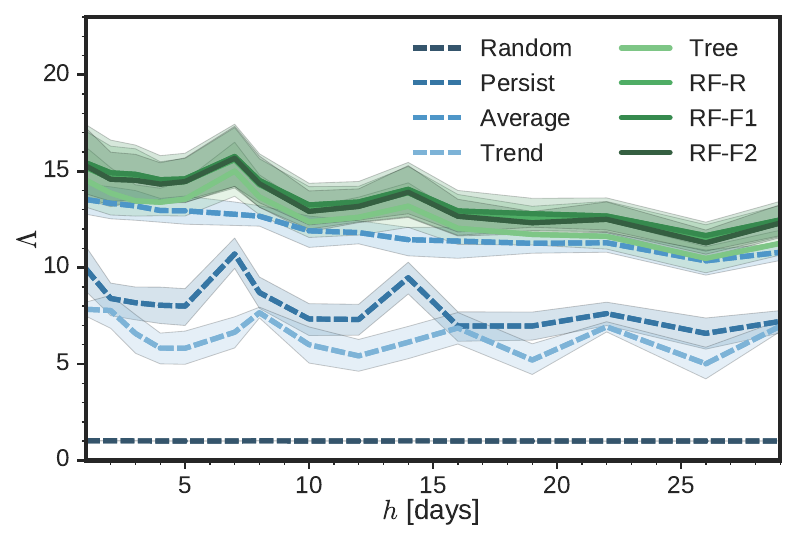}
    \vspace{\sepfigcap}
    \caption{Hot spot forecast: average lift $\Lambda$ as a function of $h$ for $w=7$. Dashed lines indicate baseline models and solid lines indicate classifier-based models. Shadowed regions indicate 95\% confidence intervals.}
    \label{fig:res_day_h}
\end{figure}

We can now focus on the performance of the models as a function of the prediction horizon $h$ (Fig.~\ref{fig:res_day_h}). First of all, we confirm that the random model has a lift $\Lambda\approx 1$, as expected. Next, we observe that the Persist and Trend models have a substantially lower performance as compared to the rest. Interestingly, for the Persist model, we observe two performance peaks at $h=7$ and $h=14$, which correspond to predicting today's label for 7 and 14 days ahead, respectively (e.g.,~if today is a Saturday, predicting the same label for next Saturday). This result stems from the weekly regularities observed in Sec.~\ref{sec:data_reg}. Another indication of weekly regularities is found in the performance peaks of the Trend model, which occur at $h=8$, $16$, $22$, and $29$~days. The fact that they occur at multiples of 7 plus one days rather than at multiples of 7 as we saw for Persist makes perfect sense: Trend forms a projection for the next day, while Persist works better as an estimation of the current day.

Notice that, four weeks ahead ($h=29$), we can still perform predictions that are more than 12 times better than random. This is an indication of the temporal stability of the data. 
For instance, if a sector has a large average score for the past week, we can infer that it has high chances of being a hot spot in the forthcoming weeks. This is clearly true for a number of sectors in our data, which are almost permanently a hot spot (cf.~Fig.~\ref{fig:basic1}C). More importantly, the reverse also applies: the majority of sectors are never or almost never a hot spot (cf.~Fig.~\ref{fig:example_y_day}). Therefore, in that case, very low scores $\ma{S}$ become a good indicator of not being a future hot spot. A manual inspection of a subset of the sectors of $\ma{S}$ further confirms that relatively high scores are typically present before becoming a hot spot. We will come back to the importance of features below.

\begin{figure}[t]
    \centering
    \includegraphics{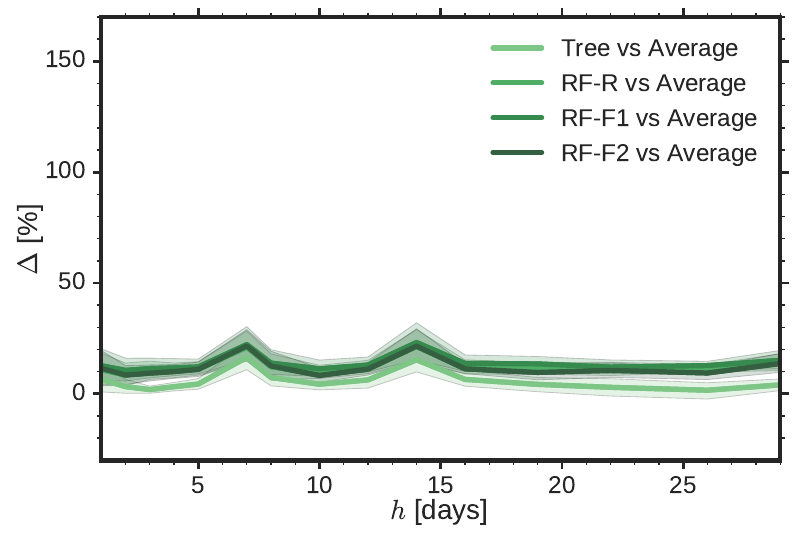}
    \vspace{\sepfigcap}
    \caption{Hot spot forecast: average ratios $\Delta$ as a function of $h$ for $w=7$. Shadowed regions indicate 95\% confidence intervals.}
    \label{fig:res_ratio_day_h}
\end{figure}

The Average model performs surprisingly well, but never reaches the performance of the classifier-based models. To see that, we can plot the comparison ratio $\Delta$ as a function of the horizon $h$ (Fig.~\ref{fig:res_ratio_day_h}). The worst classifier-based model, the Tree, is on average 6\% better than the Average model. The best classifier-based model, RF-F1, is 14\% better than the Average model. All RF models perform similarly, with ratios that go from 6\% (worst case) to 22\% (best case). Interestingly, classifier-based models also seem to exploit the weekly regularities found for the Persist model: they also exhibit the same peaks at $h=7$ and $h=14$ days. More interestingly, there seems to be a slight increase in the RF ratios for longer horizons ($h>20$). This can be due to two reasons: either the Average is not as valid as for shorter horizons, or the classifier-based models really bring some value for long horizons. We conjecture it is a mixture of both.

\begin{figure}[t]
    \centering
    \includegraphics{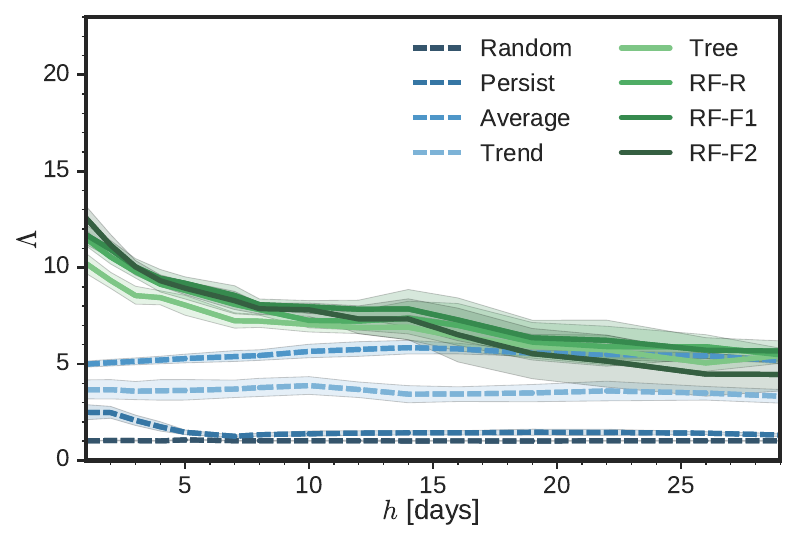}
    \vspace{\sepfigcap}
    \caption{Become a hot spot forecast: average lift $\Lambda$ as a function of $h$ for $w=7$. Dashed lines indicate baseline models and solid lines indicate classifier-based models. Shadowed regions indicate 95\% confidence intervals.}
    \label{fig:res_become_h}
\end{figure}

\begin{figure}[t]
    \centering
    \includegraphics{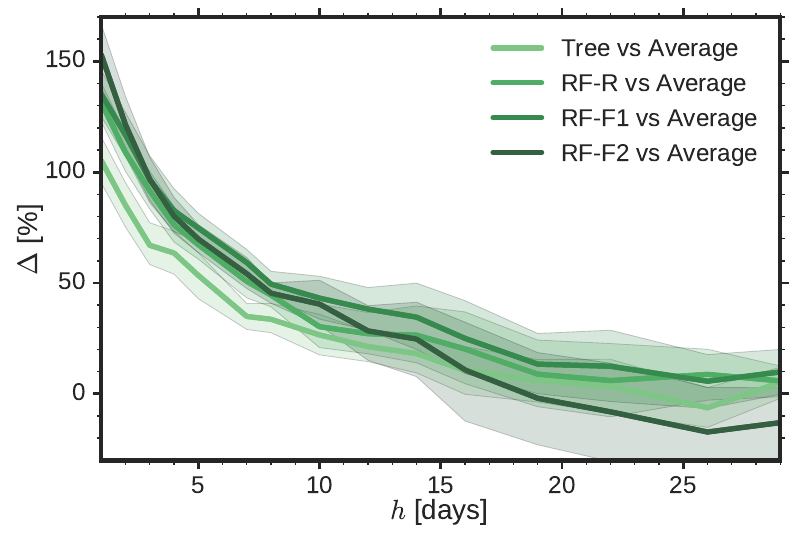}
    \vspace{\sepfigcap}
    \caption{Become a hot spot forecast: average ratios $\Delta$ as a function of $h$ for $w=7$. Shadowed regions indicate 95\% confidence intervals.}
    \label{fig:res_ratio_become_h}
\end{figure}

We now turn to the `become a hot spot' forecast, used as a contrast to the `be a hot spot' forecast considered above. With that, we switch to the task of predicting non-regular but temporally-consistent future hot spots (Fig.~\ref{fig:res_become_h}). In this new scenario, we see a clear difference in the performance of the classifier-based methods with respect to all the baselines for $h\leq 15$~days. The worst classifier-based method, again the Tree, is now 105\% better than the best baseline method, again the Average (Fig.~\ref{fig:res_ratio_become_h}). This contrasts with the previous situation (compare Figs.~\ref{fig:res_ratio_day_h} and~\ref{fig:res_ratio_become_h}).

The `become a hot spot' forecast presents almost no evidence of weekly regularities. Indeed, the aforementioned peaks around multiples of 7 are not present anymore (Fig.~\ref{fig:res_become_h}). In addition, we now see that the performance of the classifier-based models becomes comparable to the one of the Average model for $h\geq 19$~days (Fig.~\ref{fig:res_ratio_become_h}). This implies that, for long-range forecasts of non-regular hot spots, we cannot do better than a guess based on the average of the previous scores $\ma{S}$. 
This can also be observed if we inspect the first splits of the Tree model. For instance, if we consider the Tree trained for $h=22$~days, the 
score $\ma{S}$ appears already in the first split, and also in the third split, highlighting the importance of this variable to perform forecasts. 


\subsection{Past history}
\label{sec:results_past}

\begin{figure}[t]
    \centering
    \includegraphics{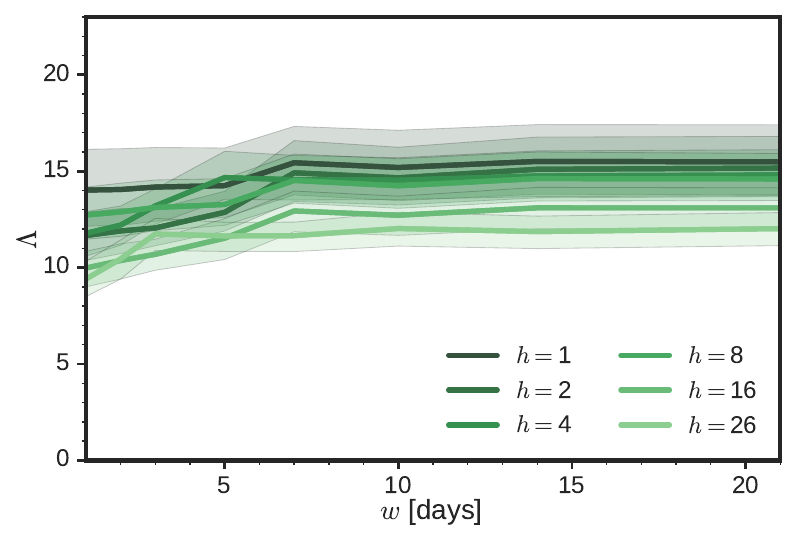}
    \vspace{\sepfigcap}
    \caption{Hot spot forecast: average lift as a function of $w$ for the RF-F1 model. Shadowed regions indicate 95\% confidence intervals. Qualitatively similar results were obtained for the other classifier-based models.}
    \label{fig:res_day_w}
\end{figure}

\begin{figure}[t]
    \centering
    \includegraphics{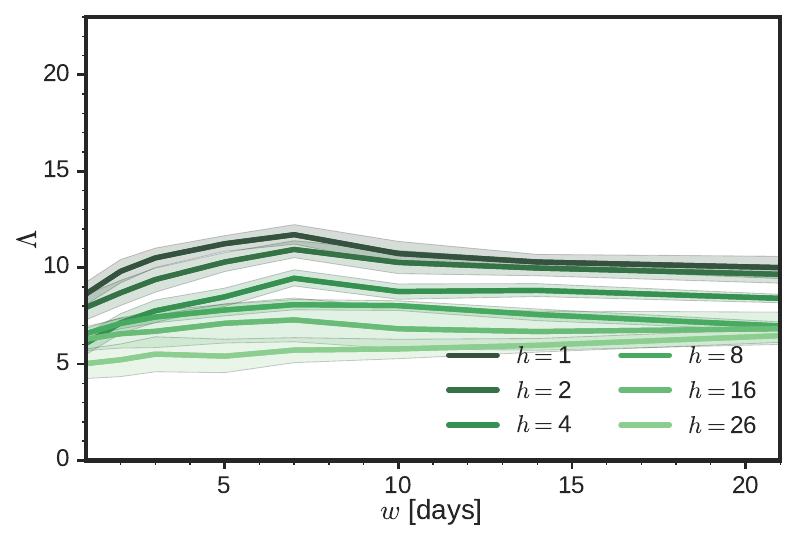}
    \vspace{\sepfigcap}
    \caption{Become a hot spot forecast: average lift as a function of $w$ for the RF-F1 model. Shadowed regions indicate 95\% confidence intervals. Qualitatively similar results were obtained for the other classifier-based models.}
    \label{fig:res_become_w}
\end{figure}

We now study the effect of the amount of past $w$ considered by the models (Fig.~\ref{fig:res_day_w}). In general, we see that performance does not change much with $w$. With only one day of past information ($w=1$), the models are already able to produce forecasts that are almost 10 times better than random. This performance is increased until $w=7$ (one week), where it reaches a plateau at the maximum lift. A similar situation occurs with the `become a hot spot' forecast (Fig.~\ref{fig:res_become_w}). However, in this case, performance slightly drops for $w>7$, and generally reaches a plateau for $w\geq 10$~days. Here, we also observe that the effect of $w$ is much less pronounced for large horizons $h$, being almost nonexistent for $h=16$ or $h=26$~days. Nonetheless, the latter shows a minimal but consistent increase with $w$. This could be an indication that, with much more data and a machine learning algorithm that can handle high dimensionalities, we could potentially improve our lift figures for long-range horizons. However, as we cannot fulfill the first condition with the current data set, we leave this aspect for future research.

\subsection{Features' importance}
\label{sec:results_features}

\begin{figure}[t]
    \centering
    \includegraphics{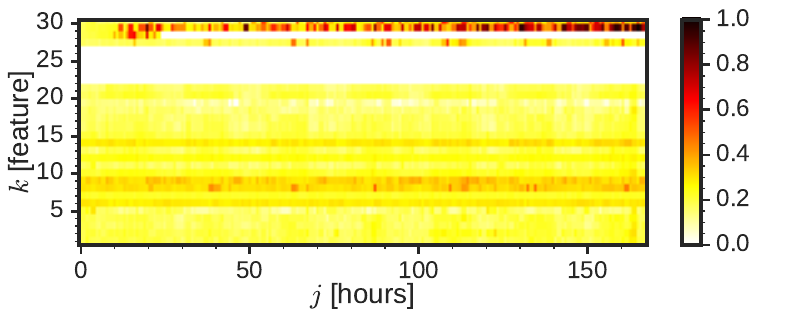}
    \vspace{\sepfigcap}
    \caption{Hot spot forecast: cumulative feature importance for the RF-R model with $h=5$ and $w=7$. Rows correspond to features' importance over time, features being sorted as in Eq.~\ref{eqn:combine}.}
    \label{fig:fi_day}
\end{figure}

We end our results section by looking at the relative importance of KPIs for the task of forecasting hot spots (Fig.~\ref{fig:fi_day}). We find that the most important feature is the weekly hotness score $\ma{S}^{\text{w}}$ ($k=29$), and that its importance increases with $j$, i.e., as we get closer to the present $t$. Isolated values of $\ma{S}^{\text{h}}$, $\ma{S}^{\text{d}}$, and $\ma{Y}^{\text{d}}$ are also taken into account ($k=27$, $28$, and $30$, respectively). Interestingly, $\ma{S}^{\text{d}}$ is only considered for time spans where the weekly score $\ma{S}^{\text{w}}$ ($k=28$) cannot integrate much temporal information ($j<25$), whereas the contribution of the hourly score $\ma{S}^{\text{h}}$ ($k=27$) is scattered along the time axis $j$, more or less at regular intervals of 24~hours. The latter implies that the model is exploiting the scores obtained at a particular daily time frame: between 15 and 18~hours. We hypothesize that this has to do with the end of the workday and commuting.

In essence, we find that forecasts are performed on the basis of past scores $\ma{S}$ (Fig.~\ref{fig:fi_day}). However, we also find KPIs having a non-negligible contribution. Presumably, this contribution is what allows the classifier-based models to improve over the Average model and the rest of the baselines. The three most important KPIs are related to usage and congestion. Two of them are related to wireless usage (number of users queuing for a high-speed channel, $k=9$, and transmission occupancy, $k=14$), and the other is related to congestion (data utilization rate, $k=8$). Interestingly, we see that congestion characteristics are mostly taken into account at a certain time frame: the same time frame found above. Finally, we find that the enhanced calendar information $\ma{C}$ ($k\in[22,26]$) has almost no contribution to the forecast.

\begin{figure}[t]
    \centering
    \includegraphics{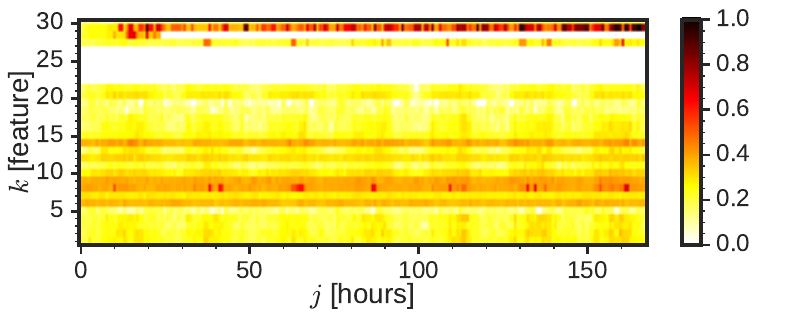}
    \vspace{\sepfigcap}
    \caption{Become a hot spot forecast: cumulative feature importance for the RF-R model with $h=5$ and $w=7$. Rows correspond to features' importance over time, features being sorted as in Eq.~\ref{eqn:combine}.}
    \label{fig:fi_become}
\end{figure}

If we consider the `become a hot spot' forecast, we can draw similar conclusions regarding $\ma{S}$ and $\ma{Y}$ (Fig.~\ref{fig:fi_become}). However, we can clearly see that, for this forecast, the importance of KPIs increases. That is, KPI information becomes more important when forecasting non-regular hot spots. Not only the aforementioned three indicators become more important, but also new ones come into play. These are related to radio interference (noise rise conditions, $k=6$, and noise absolute measurement $k=12$) and, to a minor extent, signalling (channel setup failure, $k=10$). We also see that the aforementioned data utilization rate ($k=8$) presents the same periodicity, while other highlighted indicators' importance is more or less equally spread across days and hours.

\section{Related work}
\label{sec:relatedwork}

After an exhaustive search, we have not been able to find any work that characterizes performance of hot spots in cellular networks using real data. The only exception could be the work by Nika~et~al.~\cite{hotspots}, where a data set with more than 700K users is analyzed to understand temporal characteristics of data hot spots. Predictability is assessed, but considering only hot spot repetitions at the exact time and location in the forthcoming week. However, we find that the focus of the work is not on performance, but on load, with hot spots being defined on the basis of relative daily cell traffic. 

A few further research papers have been written in an attempt to understand how load evolves in cellular networks. For instance, Paul~et~al.~\cite{dynamics} analyzed the spatio-temporal behavior of network resource usage in a real 3G network. Interestingly, their results indicate that aggregate network usage exhibits periodic behavior, but individual sectors do not exhibit such properties (however, in~\cite{correlation}, Jiang~et~al.\ do find such correlations in KPIs). The usage of individual applications in cellular data networks has been studied by Shafiq et al.~\cite{application}. Finally, simulation has been also used to quantify hot spots in wireless cellular networks~\cite{hotspots2}. None of the previous works considers real performance hot spots nor provides any forecasting methodology.

In terms of forecasting, we find that, over the past years, there have been some attempts to predict network performance based on historical network measurements. The network weather service~\cite{weather} was one of the first systems to make predictions of TCP/IP throughput and latency. Another example is the Proteus~\cite{proteus} system, which uses regression trees to forecast short-term performance in cellular networks, with the objective to proactively adapt mobile application network patterns. Similarly, Sprout~\cite{sprout} attempts to predict immediate network load in cellular networks to optimize TCP-layer parameters. Finally, in~\cite{mr}, the authors use gradient boosted trees to forecast hot spots within a data center, in order to avoid stranglers and, therefore, speed up map-reduce computation. In general, all these works focus on very short-term forecasts (in the order of seconds or, at most, minutes) and, consequently, they consider data sets spanning only very short time periods. Further approaches are either not applied to cellular networks or do not consider performance hot spots.

\section{Conclusion}
\label{sec:conclusion}

To the best of our knowledge, this is the first work to provide an in depth analysis of the dynamics of cellular network hot spots. We have done so by considering a real-world, large-scale data set of hourly KPIs measured over a period of four months, comprising tens of thousands of sectors from a top-tier cellular network operator. We have studied both temporal and spatial regularities present in our data set, and uncovered prominent hourly, daily, and weekly patterns. In addition, we have provided a formal methodology to project such patterns into the future and perform a forecast of cellular network hot spots. We have considered a variety of baseline and tree-based models, and evaluated their accuracy as a function of time, prediction horizon, and amount of past information needed. Overall, we have gained insight into the dynamics of cellular sectors and the predictability of their underperforming situations, paving the way for more proactive network operations with greater forecasting horizons.

We have conducted our evaluation using two forecasting scenarios: regular day-based hot spots, and non-regular but consistent emerging hot spots. For the former, we have showed that baseline models can capture many of the regularities present in the data, that the accuracy of all models mildly drops with the prediction horizon, and that tree-based models can outperform the best baseline by 14\%. For the latter, we have showed that the benefits of a tree-based model can be substantially larger, up to 153\% better than the best baseline, but that this improvement vanishes for prediction horizons of more than two weeks. In both scenarios, we have seen that the time of the forecast does not introduce a significant variability in the results, and that forecast accuracy reaches a plateau when at least one week of past information is considered. We have also assessed the importance of KPIs in performing such forecasts, showing that this decisively increases for the forecasting of non-regular hot spots, specially for certain usage, congestion, interference, and signalling KPIs.




\bibliographystyle{IEEEtran}
\bibliography{IEEEabrv,joan,ilias}
%
%
%

\end{document}